\documentclass[12pt, draftclsnofoot, onecolumn]{IEEEtran}
\usepackage{amsmath,amsfonts,amssymb}
\usepackage{graphicx}
\usepackage{cite}

\title{Image Segmentation using Chan-Vese Active Contours}
\author{Pranav Shenoy K. P.}%

\usepackage{listings}
\usepackage{color}
\usepackage{caption}
\usepackage{float}

\definecolor{dkgreen}{rgb}{0,0.6,0}
\definecolor{gray}{rgb}{0.5,0.5,0.5}
\definecolor{mauve}{rgb}{0.58,0,0.82}

\begin{document}

\maketitle

\begin{abstract}
This paper presents a comprehensive derivation and implementation of the Chan–Vese active contour model for image segmentation. The model, derived from the Mumford–Shah variational framework, evolves contours based on regional intensity differences rather than image gradients, making it highly effective for segmenting noisy images or images with weak boundaries. We provide a rigorous mathematical derivation of the level set formulation, including detailed treatment of each energy term using the divergence theorem and curve evolution theory. The resulting algorithm is implemented in Python using finite difference methods with special care to numerical stability, including an upwind entropy scheme and curvature-based regularization. Experimental results on medical and synthetic images demonstrate accurate segmentation, robustness to noise, and superior performance compared to classical edge-based methods. This study confirms the suitability of the Chan–Vese model for complex segmentation tasks and highlights its potential for use in real-world imaging applications.
\end{abstract}


\section{Introduction}
Image segmentation is the process of dividing the image into a set of regions. It is widely used in areas such as medical image processing to extract portions of the tissues from a larger image as well as in automotive industry for applications such as pedestrian detection and scene recognition. There are multiple approaches to segment images, they range from using classical images processing techniques such as Otsu Thresholding to edge detection methods and histogram-based methods. Chan-Vese algorithm excels in images that do not have clearly defined boundaries or edges as it is based on an energy minimization problem and does not depend on gradients or edges. In this paper, we will discuss the use of active contours to segment an image.

Active contours or ‘Snakes’ \cite{C1} are curves that move within an image with the goal of minimization of energy
to segment images. During the evolution of the curve, the goal of the active contour is to move to a lower
energy with each time step. The paper “Active Contours Without Edges” written by Tony F. Chan and
Luminita A. Vese \cite{C2}, proposes a new model for active contours to detect objects in an image based on
techniques of curve evolution, Mumford-Shah functional for segmentation and level sets.
In this paper, we will discuss about the Energy functional of Chan-Vese algorithm and derive the Gradient
descent and level set functions in Section 2. Next, we will discuss how to discretize and implement the
model on a computer in Section 3, followed by the experimental results and quantitative analysis in
Section 4. Additionally, we will also compare our results with a few other commonly used segmentation
techniques in Section 4. And finally, we will conclude by discussing what we have learnt as a result of this
project in Section 5. 

\section{The Chan-Vese Active Contour Model}
\subsection{Energy Function}
The energy functional to be minimized is defined as \cite{C3}:

\begin{equation*}
\begin{split}
F(c_1, c_2, C) =\; & \mu \cdot \text{Length}(C) + \nu \cdot \text{Area}(\text{inside}(C)) \\
& + \lambda_1 \int_{\text{inside}(C)} |u_0(x,y) - c_1|^2 \, dx\,dy + \lambda_2 \int_{\text{outside}(C)} |u_0(x,y) - c_2|^2 \, dx\,dy
\end{split}
\end{equation*}

\noindent Where $\lambda_1$, $\lambda_2$, $\mu$, and $\nu$ are the scaling parameters.

\begin{equation}
E(C,u,v) = \lambda_1 \iint_{R_{in}} (I - u)^2\,dx\,dy + \lambda_2 \iint_{R_{out}} (I - v)^2\,dx\,dy \\
+ \mu \int_C ds + \nu \iint_{R_{in}} dx\,dy
\end{equation}

\noindent\textbf{Where:}
\begin{itemize}
    \item $I$ represents the image, $u$ is the average intensity inside the contour, and $v$ is the average intensity outside the contour.
    \item $C$ is the contour, $R_{\text{in}}$ is the region inside the contour, $R_{\text{out}}$ is the region outside the contour, and $\lambda_1 > 0$, $\lambda_2 > 0$, $\mu \geq 0$, and $\nu \geq 0$ are the scaling parameters.
\end{itemize}

\noindent In Equation~(1), the first term attempts to minimize the energy inside the contour and maintain uniformity in the enclosed region. The second term attempts to minimize the energy outside the contour, also promoting uniformity. The third term penalizes the contour length, encouraging the contour to remain short and smooth, thereby avoiding unnecessary irregularities. The fourth term penalizes the area inside the contour. The scaling parameters control the influence of each term in the energy functional. As recommended by Chan and Vese, typical values are $\lambda_1 = \lambda_2 = 1$ and $\nu = 0$.

\subsection{Gradient Descent and Level Set Functions}
To derive gradient descent, we begin with:
\begin{equation}
C_t = -\nabla E
\end{equation}
We will use the following relationship to derive the gradient descent (2) of our energy functional defined
in Equation (1):
\begin{equation}
\frac{d}{dt} E(C(t)) = \left\langle \nabla E, C_t \right\rangle = \int_C (C_t \cdot \nabla E) ds
\end{equation}

\noindent This relationship states that the time derivative of the energy functional is the inner product of $\nabla E$ and $C_t$, which is also equivalent to the integral of the dot product of $\nabla E$ and $C_t$ over the contour $C$. As a result, the time derivative of the energy functional can be simplified to this form, which allows us to extract $\nabla E$ once $C_t$ is isolated. Once these terms are obtained, they can be used to compute the level set function $\psi_t$.

\noindent Since the integrand of Equation~(1) is large, we compute the level set function contributions for each term individually and combine them afterward.

\subsection*{\textbf{Term 1: $\iint (I - u)^2 \, dx\,dy$}}

\vspace{0.5em}

\noindent For simplicity, let us define $f_{\text{in}} = (I - u)^2$. 
From Equation~(3), we obtain $f_{\text{in}}$ integrated over the region inside the contour:

Let $f_{in} = (I - u)^2$. Then:
\begin{equation}
E(C(t)) = \iint_{R_{in}} f_{in} \, dx \, dy
\end{equation}
To simplify the equation, we can rewrite the area integral as a contour integral. For this, we can use the
Divergence Theorem:
\begin{equation}
\int_C \vec{F} \cdot \vec{N} \, ds = \iint_R \nabla \cdot \vec{F} \, dx\,dy
\end{equation}

We can rewrite the energy functional (4) in terms of a contour integral if we set $f_{\text{in}} = \nabla \cdot \vec{F}$. Then we get:

\begin{equation}
E(C(t)) = \iint_{\text{R}_{\text{in}}} f_{\text{in}} \, dx\,dy = \iint_{\text{R}_{\text{in}}} (\nabla \cdot \vec{F}) \, dx\,dy = \int_{C} \vec{F} \cdot \vec{N} \, ds
\end{equation}

Here, $\vec{F} = [F_1 \;\; F_2]^T$ and $\nabla \cdot \vec{F} = \frac{\partial F_1}{\partial x} + \frac{\partial F_2}{\partial y}$

Next, let us differentiate both sides with respect to $t$:
\begin{equation}
\frac{d}{dt} E(C(t)) = \frac{d}{dt} \int_C \vec{F} \cdot \vec{N} \, ds
\end{equation}

To ‘bring’ the differential inside the integral in RHS (right hand side), we need to replace $ds$ with $dp$. The relationship between them is given by: $\|\mathcal{C}_p\| dp = ds$. While we are replacing, we must also remember to change the integral over $C$ to integral over 1 to 0.

\begin{equation}
\frac{d}{dt} E(C(t)) = \frac{d}{dt} \int_0^1 \vec{F} \cdot \vec{N} \, \|\mathcal{C}_p\| \, dp
\end{equation}

Next, we can differentiate the integrand using Chain Rule:
\begin{equation}
\frac{dE}{dt} = \int_0^1 \left( \frac{d}{dt} (\vec{F}) \cdot \vec{N} \, \|\mathcal{C}_p\| \right) + \left( \vec{F} \cdot \frac{d}{dt} (\vec{N}) \, \|\mathcal{C}_p\| \right) + \left( \vec{F} \cdot \vec{N} \, \frac{d}{dt} (\|\mathcal{C}_p\|) \right) \, dp
\end{equation}

For simplicity, let us divide and simplify the 3 terms independently and combine it later.

\subsection*{Term 1: $\int_0^1 \left( \frac{d}{dt} (\vec{F}) \cdot \vec{N} \|\mathcal{C}_p\| \right) dp$}

\noindent
$\dfrac{d}{dt} (\vec{F})$ can be written as 
$\left[ \dfrac{\partial \vec{F}}{\partial X} \right] C_t$. 
Where 
\[
\left[ \frac{\partial \vec{F}}{\partial X} \right] =
\begin{bmatrix}
\frac{\partial \vec{F}_1}{\partial x} & \frac{\partial \vec{F}_1}{\partial y} \\
\frac{\partial \vec{F}_2}{\partial x} & \frac{\partial \vec{F}_2}{\partial y}
\end{bmatrix}.
\]

\noindent
Therefore:
\[
\int_0^1 \left( \frac{d}{dt} (\vec{F}) \cdot \vec{N} \|\mathcal{C}_p\| \right) dp
=
\int_0^1 \left( \left[ \frac{\partial \vec{F}}{\partial X} \right] C_t \cdot \vec{N} \|\mathcal{C}_p\| \right) dp
\tag{10}
\]

\vspace{1em}

\noindent
Let us now switch back to $ds$ from $dp$:

\[
\int_0^1 \left( \frac{d}{dt} (\vec{F}) \cdot \vec{N} \|\mathcal{C}_p\| \right) dp 
= \int_0^1 \left( \left[ \frac{\partial \vec{F}}{\partial X} \right] C_t \cdot \vec{N} \right) \|\mathcal{C}_p\| dp 
= \int_C \left[ \frac{\partial \vec{F}}{\partial X} \right] C_t \cdot \vec{N} \, ds 
\tag{11}
\]

\subsection*{Term 2: $\int_0^1 \left( \vec{F} \cdot \frac{d}{dt} (\vec{N}) \|\mathcal{C}_p\| \right) dp$}

\noindent
We know that $\vec{N} = J \vec{T}^\perp$. Where $J = \begin{bmatrix} 0 & 1 \\ -1 & 0 \end{bmatrix}$ is a rotational matrix. Also, $J \vec{N} = -\vec{T}$. And $\vec{T} = \dfrac{C_p}{\|C_p\|}$.

\[
\frac{d}{dt}(\vec{N}) = \frac{d}{dt} \left( \frac{J C_p}{\|C_p\|} \right)
\]
\[
\frac{d}{dt}(\vec{N}) = J \frac{d}{dt} \left( \frac{C_p}{\|C_p\|} \right)
\]
\[
\frac{d}{dt}(N_t) = \frac{J}{\|C_p\|} \left( C_{pt} - \vec{T}(C_{pt} \cdot \vec{T}) \right)
\]
\[
N_t = \frac{J}{\|C_p\|} \left( \vec{N} \, C_{pt} \cdot \vec{N} \right)
\quad \Rightarrow \quad
N_t = \frac{-\vec{T}(C_{pt} \cdot \vec{N})}{\|C_p\|}
\]

\[
\int_0^1 \left( \vec{F} \cdot \frac{d}{dt}(\vec{N}) \|\mathcal{C}_p\| \right) dp 
= - \int_0^1 \vec{F} \cdot \frac{\vec{T}(C_{pt} \cdot \vec{N})}{\|C_p\|} \cdot \|\mathcal{C}_p\| \, dp 
\tag{12}
\]

\noindent
We can switch the order of differentiation from $C_{pt}$ to $C_{tp}$. Also, we can change $C_{tp} = C_{ts} \cdot \|C_p\|$. Let us now switch back to $ds$ from $dp$:

\[
\int_0^1 \left( \vec{F} \cdot \frac{d}{dt}(\vec{N}) \|\mathcal{C}_p\| \right) dp 
= - \int_0^1 \vec{F} \cdot \frac{\vec{T}(C_{tp} \cdot \vec{N})}{\|C_p\|} \cdot \|\mathcal{C}_p\| \, dp 
= - \int_C \vec{F} \cdot \vec{T}(C_{ts} \cdot \vec{N}) \, ds 
\tag{13}
\]

\subsection*{Term 3: $\int_0^1 \left( \vec{F} \cdot \vec{N} \cdot \frac{d}{dt}(\|\mathcal{C}_p\|) \right) dp$}

\[
\frac{d}{dt}(\|\mathcal{C}_p\|) \, dp = \frac{d}{dt} \sqrt{C_p \cdot C_p} = \frac{C_{pt} \cdot C_p}{\|C_p\|}
\]

\[
\int_0^1 \left( \vec{F} \cdot \vec{N} \frac{d}{dt}(\|\mathcal{C}_p\|) \right) dp 
= \int_0^1 \vec{F} \cdot \vec{N} \left( \frac{C_{pt} \cdot C_p}{\|C_p\|} \right) dp 
\tag{14}
\]

\noindent
We can switch the order of differentiation from $C_{pt}$ to $C_{tp}$. Also, we can change $C_{tp} = C_{ts} \cdot \|C_p\|$. Let us now switch back to $ds$ from $dp$:

\[
\int_0^1 \left( \vec{F} \cdot \vec{N} \frac{d}{dt}(\|\mathcal{C}_p\|) \right) dp 
= \int_0^1 \vec{F} \cdot \vec{N} \left( \frac{C_{tp} \cdot C_p}{\|C_p\|} \right) dp 
= \int_C \vec{F} \cdot \vec{N} (C_{ts} \cdot \vec{T}) \, ds 
\tag{15}
\]

\noindent
Let us combine the 3 terms from Equations (11), (13), (15):

\[
\frac{dE}{dt} = \int_C \left( \left[ \frac{\partial \vec{F}}{\partial X} \right] C_t \cdot \vec{N} \right)
- \left( \vec{F} \cdot \vec{T}(C_{ts} \cdot \vec{N}) \right)
+ \left( \vec{F} \cdot \vec{N}(C_{ts} \cdot \vec{T}) \right) ds
\tag{16}
\]

\noindent
Applying integration by parts, we get:

\[
\frac{dE}{dt} = \int_C \left( \left[ \frac{\partial \vec{F}}{\partial X} \right] C_t \cdot \vec{N} \right)
+ C_t \left( (\vec{F} \cdot \vec{T}) \vec{N} \right)_s
- C_t \left( (\vec{F} \cdot \vec{N}) \vec{T} \right)_s \, ds
\tag{17}
\]

\[
\frac{dE}{dt} = \int_C C_t \left( \left( \left[ \frac{\partial \vec{F}}{\partial X} \right]^T \vec{N} \right)
+ \left( (\vec{F} \cdot \vec{T}) \vec{N} \right)_s
- \left( (\vec{F} \cdot \vec{N}) \vec{T} \right)_s \right) ds
\tag{18}
\]

\noindent
This is of the form: $\int_C (C_t \cdot \nabla_C E) ds$. Therefore,

\[
\nabla_C E = \left( \left[ \frac{\partial \vec{F}}{\partial X} \right]^T \vec{N} \right)
+ \left( (\vec{F} \cdot \vec{T}) \vec{N} \right)_s
- \left( (\vec{F} \cdot \vec{N}) \vec{T} \right)_s
\]

\noindent
If we expand the terms by differentiating, we get:

\[
\nabla_C E = \left[ \frac{\partial \vec{F}}{\partial X} \right]^T \vec{N}
+ (\vec{F}_s \cdot \vec{T}) \vec{N} 
+ (\vec{F} \cdot \vec{T}_s) \vec{N}
+ (\vec{F} \cdot \vec{T}) \vec{N}_s 
- (\vec{F}_s \cdot \vec{N}) \vec{T} 
- (\vec{F} \cdot \vec{N}_s) \vec{T} 
- (\vec{F} \cdot \vec{N}) \vec{T}_s 
\tag{19}
\]

\[
\vec{F}_s = \frac{\partial}{\partial x} \vec{F}(C(s, t)) 
= \left[ \frac{\partial \vec{F}}{\partial X} \right] \vec{T}
\]

\[
\vec{T}_s = -K \vec{N}, \quad \vec{N}_s = K \vec{T}
\]

\noindent
Where $K$ is the curvature of the contour.

Using the 3 equations mentioned above, we can rewrite Equation (19) as:
\begin{align}
\nabla_C E =\; & \left[ \frac{\partial \vec{F}}{\partial X} \right]^T \vec{N} 
+ \left( \vec{T}^T \left[ \frac{\partial \vec{F}}{\partial X} \right] \vec{N} \right) \vec{T} 
- K (\vec{F} \cdot \vec{N}) \vec{N} \nonumber \\
& + K (\vec{F} \cdot \vec{T}) \vec{T} 
- \left( \vec{N}^T \left[ \frac{\partial \vec{F}}{\partial X} \right] \vec{T} \right) \vec{T} 
- K (\vec{F} \cdot \vec{T}) \vec{T} 
+ K (\vec{F} \cdot \vec{N}) \vec{N}
\tag{19}
\end{align}

We can decompose the first term of RHS into its tangential and normal components as follows:
\[
\left[ \frac{\partial \vec{F}}{\partial X} \right]^T \vec{N} = 
\left( \vec{T}^T \left[ \frac{\partial \vec{F}}{\partial X} \right] \vec{N} \right) \vec{T} 
+ \left( \vec{N}^T \left[ \frac{\partial \vec{F}}{\partial X} \right] \vec{N} \right) \vec{N}
\]

Therefore, we can rewrite Equation (19) as:
\[
\nabla_C E = \left( \vec{T}^T \left[ \frac{\partial \vec{F}}{\partial X} \right] \vec{N} \right) \vec{T} 
+ \left( \vec{N}^T \left[ \frac{\partial \vec{F}}{\partial X} \right] \vec{N} \right) \vec{N} 
+ \left( \vec{T}^T \left[ \frac{\partial \vec{F}}{\partial X} \right] \vec{N} \right) \vec{N} 
- \left( \vec{N}^T \left[ \frac{\partial \vec{F}}{\partial X} \right] \vec{T} \right) \vec{T}
\tag{20}
\]

\[
\nabla_C E = \left( \vec{N}^T \left[ \frac{\partial \vec{F}}{\partial X} \right] \vec{N} \right) \vec{N} 
+ \left( \vec{T}^T \left[ \frac{\partial \vec{F}}{\partial X} \right] \vec{N} \right) \vec{N}
\tag{21}
\]

It can be shown that $\vec{N}$ and $\vec{T}$ are orthonormal basis since these vectors are perpendicular to each other.

Also $\left[ \frac{\partial \vec{F}}{\partial X} \right]$ can be diagonalizable. Therefore, we can use the Similarity Transform which states that $a^T A a + b^T A b = \text{trace}(A)$, if $a$ and $b$ are orthonormal basis and $A$ is diagonalizable.

\[
\nabla_C E = \text{trace} \left( \left[ \frac{\partial \vec{F}}{\partial X} \right] \right) \vec{N} = 
\text{trace} \left(
\begin{bmatrix}
\frac{\partial F_1}{\partial x} & \frac{\partial F_1}{\partial y} \\
\frac{\partial F_2}{\partial x} & \frac{\partial F_2}{\partial y}
\end{bmatrix}
\right) \vec{N}
\quad \text{where } \nabla \cdot \vec{F} = f_{\text{in}}, \quad \nabla \cdot \vec{F} \cdot \vec{N} = f_{\text{in}} \vec{N}
\tag{22}
\]

With the result in (22), we can now define our gradient descent function:
\[
\boxed{
C_t = -\nabla_C E = -f_{\text{in}} \vec{N}
}
\tag{23}
\]

Now that we have $C_t$, we can compute $\Psi_t$. The relationship between curve evolution and level set evolution is given by:
\[
\Psi_t = -\nabla \Psi \cdot C_t
\]

Substituting $C_t$, we get:
\[
\Psi_t = \nabla \Psi \cdot f_{\text{in}} \vec{N}
\tag{24}
\]

\[
\vec{N} = \frac{\nabla \Psi}{\|\nabla \Psi\|}
\]

Also,
\[
\nabla \Psi = \|\nabla \Psi\| \cdot \frac{\nabla \Psi}{\|\nabla \Psi\|}
\]

Hence,
\[
\boxed{
\Psi_t = f_{\text{in}} \|\nabla \Psi\|
}
\tag{25}
\]

\subsection*{Term 2: $\iint_{R_{\text{out}}} (I - v)^2 dx\,dy$}

This term is similar to the first term except that the integral is over the region outside the contour. We can show that:

\[
\iint_{R_{\text{out}}} f_{\text{out}} dx\,dy 
= \iint_{\Omega} f_{\text{out}} dx\,dy - \iint_{R_{\text{in}}} f_{\text{out}} dx\,dy
\tag{5}
\]

The first term in RHS does not depend on the curve. Hence the gradient of this with respect to the curve is zero. Therefore,
\[
\iint_{R_{\text{out}}} f_{\text{out}} dx\,dy 
= - \iint_{R_{\text{in}}} f_{\text{out}} dx\,dy
\tag{27}
\]

Using the results from term 1, we can show that
\[
\boxed{
C_t = -\nabla E = f_{\text{out}} \vec{N}
}
\tag{28}
\]

And hence,
\[
\boxed{
\Psi_t = -f_{\text{out}} \|\nabla \Psi\|
}
\tag{29}
\]

\subsection*{Term 3: $\int_C ds$}

We can solve for $C_t$ and $\Psi_t$ using the same techniques as we did for terms 1 and 2.

\[
E(C(t)) = \int_C ds
\]

Differentiating both sides with respect to $t$:

\[
\frac{d}{dt} E(C(t)) = \frac{d}{dt} \int_C ds = \frac{d}{dt} \int_0^1 \|\mathcal{C}_p\| dp = \int_0^1 \|\mathcal{C}_p\|_t dp
\tag{30}
\]

Now, we can use substitute $\|\mathcal{C}_p\|_t = \left( \sqrt{C_p \cdot C_p} \right)_t = \frac{C_{pt} \cdot C_p}{\|C_p\|}$, simplify, and swap $t$ and $p$ derivative parameters:

\[
\frac{dE}{dt} = \int_0^1 \left( \frac{C_{pt} \cdot C_p}{\|C_p\|} \right) dp = \int_0^1 (C_{pt} \cdot \vec{T}) dp = \int_0^1 (C_{tp} \cdot \vec{T}) dp
\tag{6}
\]

By applying integration by parts we get:

\[
\frac{dE}{dt} = -\int_0^1 (C_t \cdot \vec{T}_p) dp 
= -\int_0^1 (C_t \cdot \vec{T}_s) \|\mathcal{C}_p\| dp 
= -\int_0^1 (C_t \cdot \vec{T}_s) ds
\tag{32}
\]

We know that $\vec{T}_s = K \vec{N}$, where $K$ is the curvature of the contour. Therefore:

\[
\frac{dE}{dt} = \int_0^1 C_t (K \vec{N}) ds
\tag{33}
\]

\noindent
This is of the form $\int_C (C_t \cdot \nabla_C E) ds$. Therefore,

\[
\boxed{
C_t = -\nabla E = K \vec{N}
}
\tag{34}
\]

\noindent
The value of curvature $K = -\nabla \cdot \left( \frac{\nabla \Psi}{\|\nabla \Psi\|} \right)$, therefore the level-set function is:

\[
\boxed{
\Psi_t = -K \|\nabla \Psi\| = \nabla \cdot \left( \frac{\nabla \Psi}{\|\nabla \Psi\|} \right) \|\nabla \Psi\|
}
\tag{35}
\]

\subsection*{Term 4: $\iint_{R_{\text{in}}} dx\,dy$}

This term looks similar to the first term if $f_{\text{in}} = 1$. Hence, we can compute the $C_t$ and $\Psi_t$ by considering $f_{\text{in}} = 1$.

\[
\boxed{
C_t = -\nabla E = -\vec{N}
}
\tag{36}
\]

\[
\boxed{
\Psi_t = \|\nabla \Psi\|
}
\tag{37}
\]

Finally, by combining the level-set functions found in Equation (35) on the original energy functional in \textbf{(1)}, the level-set gradient descent equation for the Chan-Vese algorithm is computed as:

\[
\boxed{
\Psi_t = \lambda_1 (I - u)^2 \|\nabla \Psi\| - \lambda_2 (I - v)^2 \|\nabla \Psi\| 
+ \mu \nabla \cdot \left( \frac{\nabla \Psi}{\|\nabla \Psi\|} \right) \|\nabla \Psi\| + \nu \|\nabla \Psi\|
}
\tag{38}
\]

We can simplify the third term of the RHS using first and second order derivatives as follows:

\[
\mu \nabla \cdot \left( \frac{\nabla \Psi}{\|\nabla \Psi\|} \right) \|\nabla \Psi\| 
= \mu \cdot \frac{\Psi_{yy} \Psi_x^2 - 2\Psi_x \Psi_y \Psi_{xy} + \Psi_{xx} \Psi_y^2}{\Psi_x^2 + \Psi_y^2}
\]

\subsection{Upwind Entropy Norm}
Due to non-linear terms, we cannot use central differences for the approximation for $\|\nabla \Psi\|$, hence we must use forward and backward differences. The upwind entropy scheme is as follows:

\[
a \|\nabla \Psi\| =
\begin{cases}
a \cdot \sqrt{ \max^2(D_x^+) + \min^2(D_x^-) + \max^2(D_y^+) + \min^2(D_y^-) }, & a > 0 \\
a \cdot \sqrt{ \min^2(D_x^+) + \max^2(D_x^-) + \min^2(D_y^+) + \max^2(D_y^-) }, & a < 0
\end{cases}
\tag{39}
\]

Hence based on the value of $a$, we must use a combination for forward and backward differences as given above.

\section{Discretization and Implementation}
\subsection{Discretization}
For most of the discretization, we can use the Central Difference Scheme since they involve first or second
order derivatives. The central difference equations that are used to approximate are as follows:
\begin{align*}
\psi_x(x, y, t) &= \frac{\psi(x+\Delta x, y, t) - \psi(x-\Delta x, y, t)}{2\Delta x} \\
\psi_y(x, y, t) &= \frac{\psi(x, y+\Delta y, t) - \psi(x, y-\Delta y, t)}{2\Delta y} \\
\psi_{xx}(x, y, t) &= \frac{\psi(x+\Delta x, y, t) - 2\psi(x, y, t) + \psi(x-\Delta x, y, t)}{\Delta x^2} \\
\psi_{yy}(x, y, t) &= \frac{\psi(x, y+\Delta y, t) - 2\psi(x, y, t) + \psi(x, y-\Delta y, t)}{\Delta y^2} \\
\psi_{xy}(x, y, t) &= \frac{\psi_x(x, y+\Delta y, t) - \psi_x(x, y-\Delta y, t)}{2\Delta y}
\end{align*}

CFL conditions for stability:
\begin{equation*}
\Delta t \leq \frac{\Delta x^2}{2}, \quad \Delta t \leq \frac{\Delta y^2}{2}
\tag{40}
\end{equation*}

Finally, the backward and forward differences used for the upwind entropy norm can be approximated
with the following equations. To avoid oscillations in nonlinear terms, forward and backward differences are used:

\begin{align*}
D_x^+(x, y, t) &= \frac{\psi(x+\Delta x, y, t) - \psi(x, y, t)}{\Delta x}, \\
D_x^-(x, y, t) &= \frac{\psi(x, y, t) - \psi(x-\Delta x, y, t)}{\Delta x}, \\
D_y^+(x, y, t) &= \frac{\psi(x, y+\Delta y, t) - \psi(x, y, t)}{\Delta y}, \\
D_y^-(x, y, t) &= \frac{\psi(x, y, t) - \psi(x, y-\Delta y, t)}{\Delta y}
\tag{41}
\end{align*}

\section{Experiments and Results}

We implemented the Chan–Vese algorithm in Python, based on the variational model described in the previous sections. To evaluate its effectiveness, we applied the method to a variety of real-world images, including medical images \cite{C4}.

\begin{figure}[ht]
    \centering
    \includegraphics[width=0.9\textwidth]{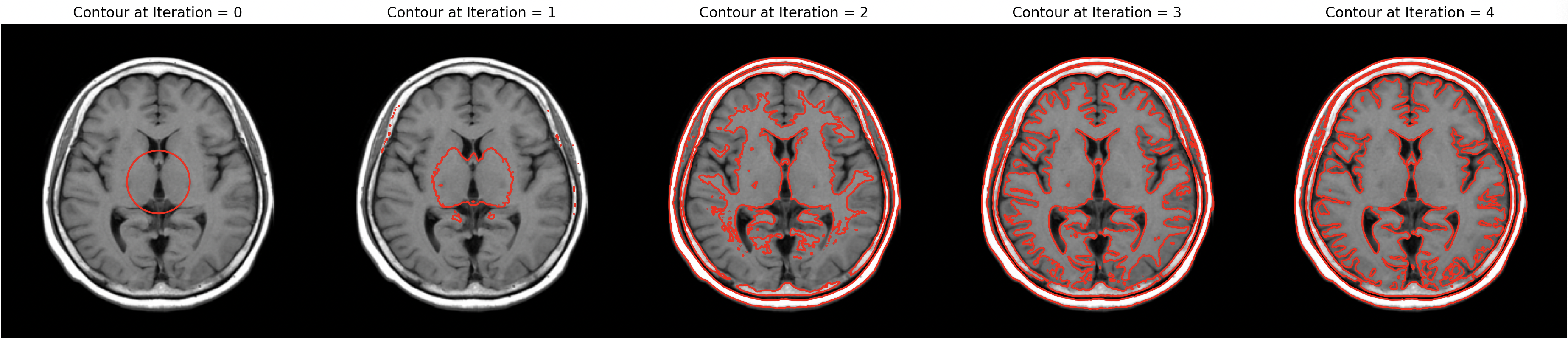}
    \caption{Evolution of contours over iterations}
    \label{fig:contour}
\end{figure}

\begin{figure}[ht]
    \centering
    \includegraphics[width=0.9\textwidth]{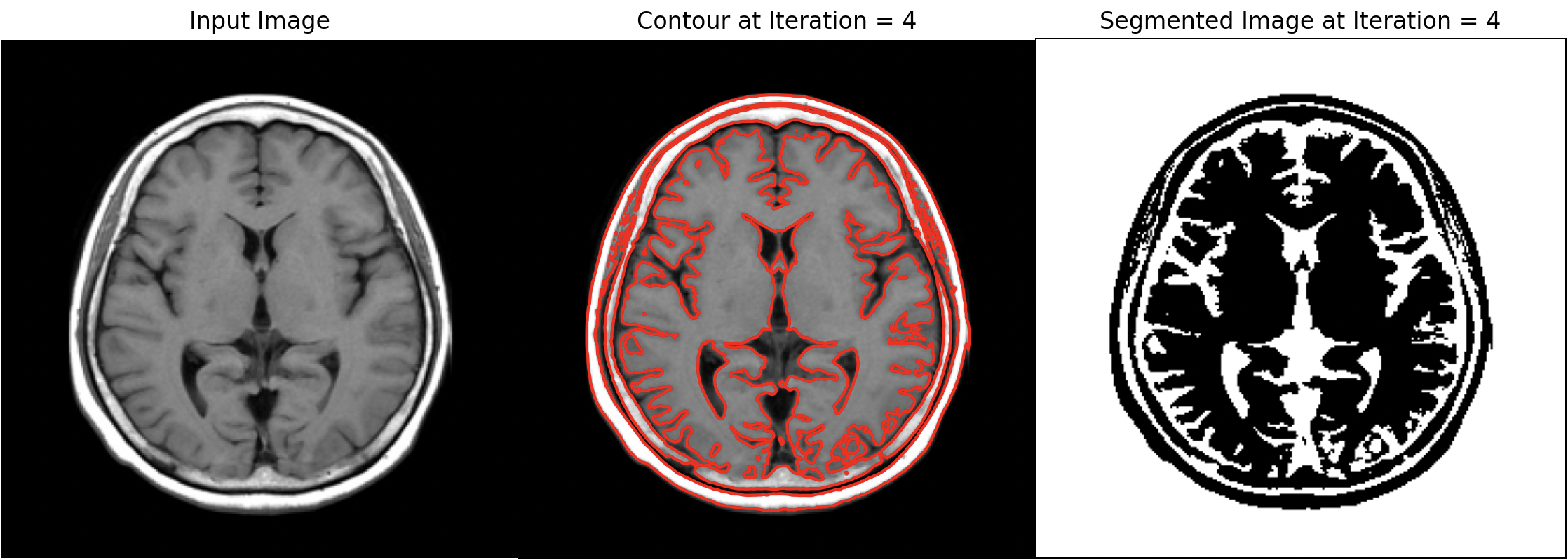}
    \caption{Segmentation of brain tissue}
    \label{fig:segmentation}
\end{figure}

\begin{figure}[ht]
    \centering
    \includegraphics[width=0.8\textwidth]{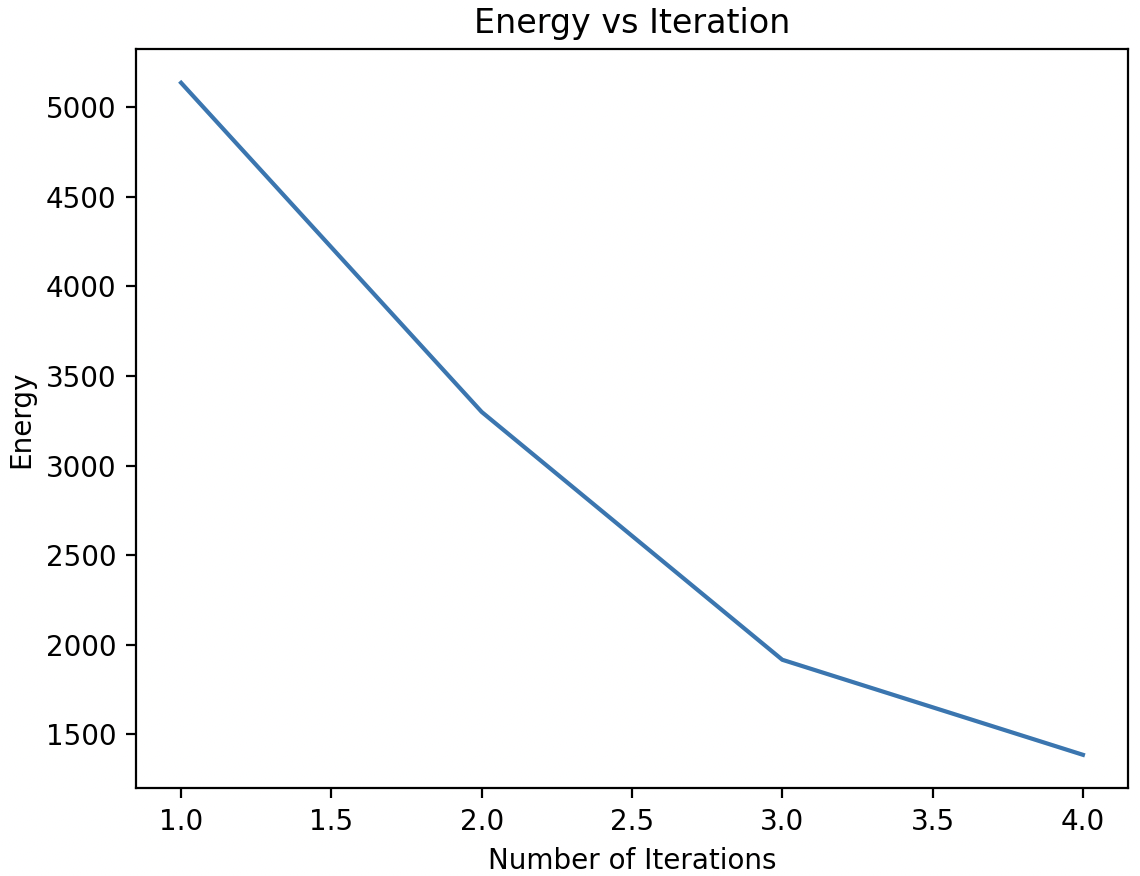}
    \caption{Plot of energy and iterations}
    \label{fig:energy}
\end{figure}

Figures 1 and 2 illustrate segmentation contours on a sample MRI brain scan image. Here the evolving contour accurately delineates object boundaries despite fuzziness and weak gradients. While Figure 3 shows the minimization of energy over iterations.

\begin{figure}[ht]
    \centering
    \includegraphics[width=1.0\textwidth]{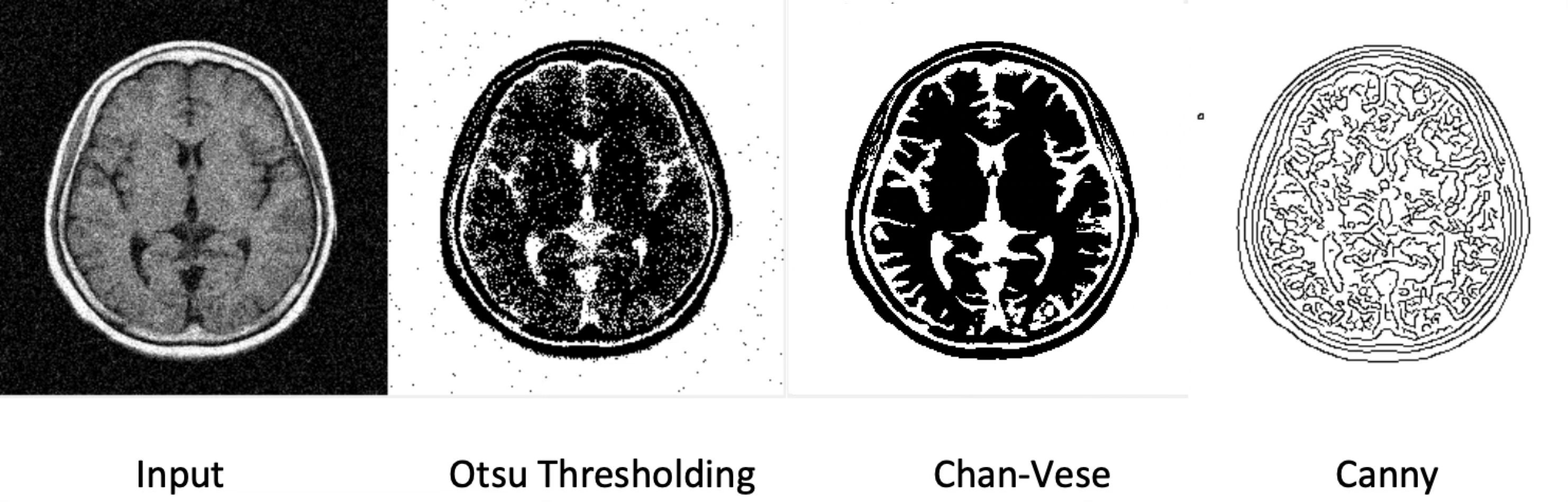}
    \caption{Comparison with edge and gradient based methods}
    \label{fig:energy}
\end{figure}

In Figure 4, we can see that the Chan–Vese model does not depend on edge information, allowing it to successfully segment objects in fuzzy or noisy images.

These experiments demonstrate the effectiveness of the algorithm in segmenting objects with poorly defined boundaries. Unlike edge-based level set methods using image gradient, region-based methods usually utilize the global region information to stabilize their responses to local variations (such as weak boundaries and noise). Thus, they can obtain a better performance of segmentation than edge-based level set methods, especially for images with weak object boundaries and noise. \cite{C5}.

\section{Algorithmic Enhancements to the Classical Chan-Vese Implementation}
In this section, we summarize key improvements over the classical Chan–Vese algorithm \cite{C6} \cite{C7} \cite{C8}. This implementation introduces practical enhancements that improve efficiency, stability, and usability.

The implementation includes several enhancements to improve efficiency and stability. Curvature is computed only in a narrow band around the contour, reducing computation without sacrificing accuracy. An upwind entropy scheme replaces central differences for $|\nabla \psi|$, improving numerical stability. The level-set function $\phi$ is initialized using a signed distance transform (bwdist) for a more accurate starting contour. Convergence is assessed through logical XOR of masks and periodic Sussman reinitialization \cite{C9} maintains the signed distance property. Key components such as SDF computation, curvature, convergence checks, and visualization are modularized to support easier debugging and future extensions.

These improvements enhance contour stability and robustness to noise, making the method effective for segmenting point clusters, as shown below.

\begin{figure}[H]
    \centering
    \includegraphics[width=1.0\textwidth]{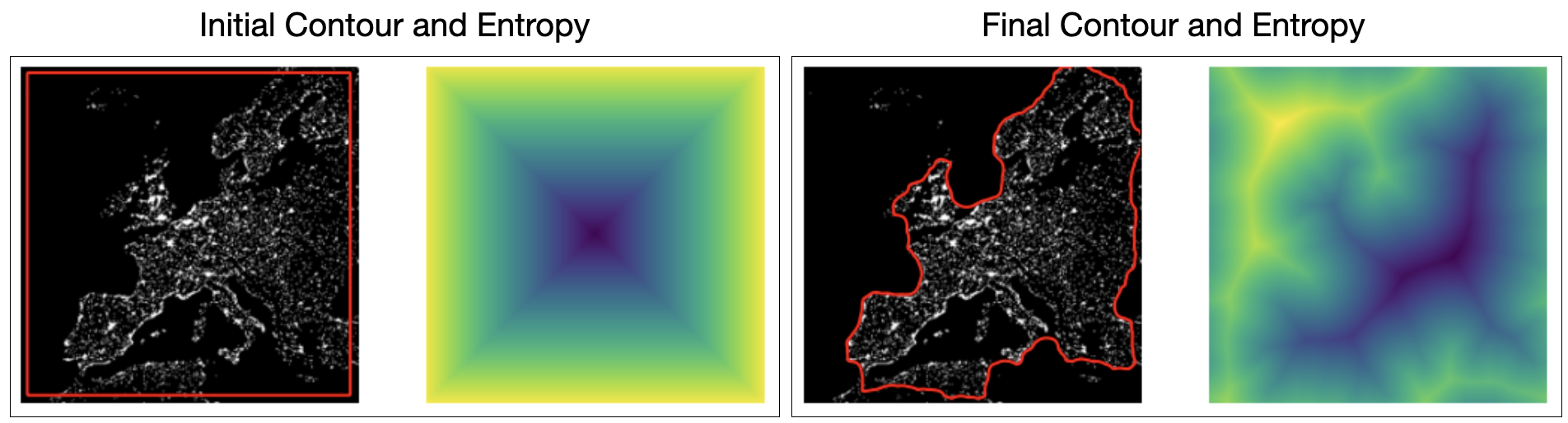}
    \label{fig:europe}
    \vspace{-5em}
\end{figure}

\begin{figure}[H]
    \centering
    \includegraphics[width=1.0\textwidth]{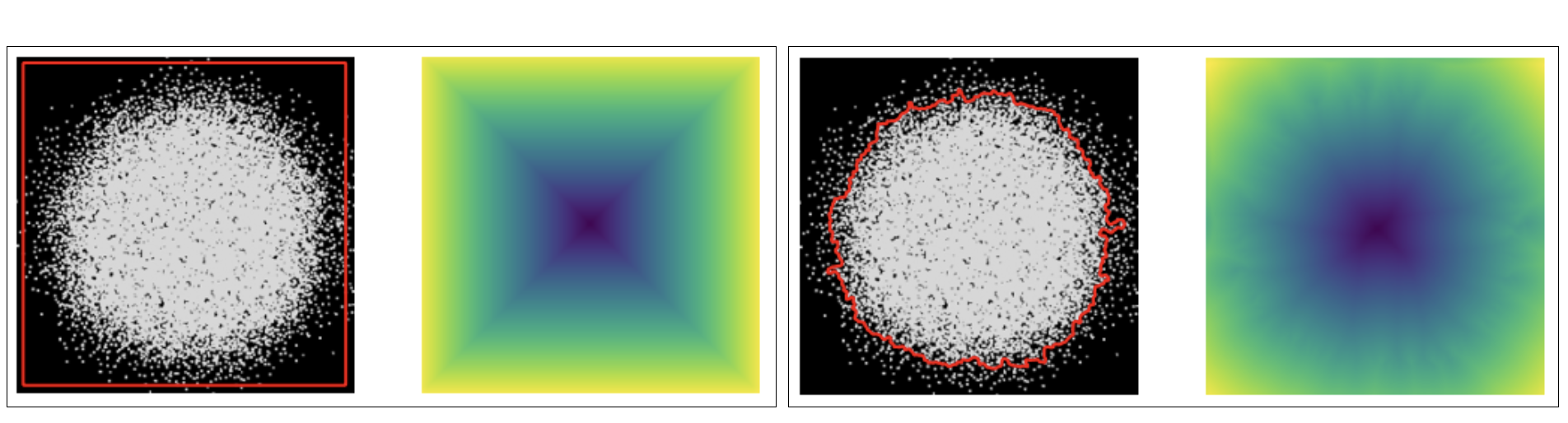}
    \captionsetup{justification=centering}
    \caption{Chan-Vese Contours on dot clusters}
    \label{fig:dots}
\end{figure}

\section{Conclusion}
In this paper, we derived the energy function for the Chan-Vese algorithm and used it to segment images. This approach proves effective in segmenting noisy images and those with weak gradients, as it relies on energy minimization to draw contours rather than edge-based methods.



\bibliographystyle{IEEEtran}  

\section*{Appendix}
\begin{lstlisting}

// Basic Chan-Vese Algorithm
import numpy as np
import matplotlib.pyplot as plt
from scipy.ndimage import gaussian_filter
from PIL import Image

def curvature(phi):
    Phi_x = (np.roll(phi, -1, axis=0) - np.roll(phi, 1, axis=0)) / 2.0
    Phi_y = (np.roll(phi, -1, axis=1) - np.roll(phi, 1, axis=1)) / 2.0
    Phi_xy = (np.roll(Phi_x, -1, axis=1) - np.roll(Phi_x, 1, axis=1)) / 2.0
    Phi_xx = np.roll(phi, -1, axis=0) - 2 * phi + np.roll(phi, 1, axis=0)
    Phi_yy = np.roll(phi, -1, axis=1) - 2 * phi + np.roll(phi, 1, axis=1)
    denom = np.maximum(np.finfo(float).eps, Phi_x**2 + Phi_y**2)
    K = -(Phi_xx * Phi_y**2 - 2 * Phi_x * Phi_y * Phi_xy + Phi_yy * Phi_x**2) / denom
    return K

def upwind_entropy_norm(phi, sign):
    eps = np.finfo(float).eps

    DxF = np.roll(phi, -1, axis=0) - phi
    DxB = phi - np.roll(phi, 1, axis=0)
    DyF = np.roll(phi, -1, axis=1) - phi
    DyB = phi - np.roll(phi, 1, axis=1)

    if sign == 1:
        norm = np.sqrt(
            np.sum(
                (np.maximum(DxF, eps))**2 +
                (np.minimum(DxB, eps))**2 +
                (np.maximum(DyF, eps))**2 +
                (np.minimum(DyB, eps))**2
            )
        )
    elif sign == -1:
        norm = np.sqrt(
            np.sum(
                (np.minimum(DxF, eps))**2 +
                (np.maximum(DxB, eps))**2 +
                (np.minimum(DyF, eps))**2 +
                (np.maximum(DyB, eps))**2
            )
        )
    else:
        raise ValueError("Sign must be 1 or -1")

    return norm

def chan_vese(I, phi0, tau, maxT, lambda1, lambda2, mu, nu):
    niter = int(round(maxT / tau))
    niter = 10
    E = np.zeros(niter)
    phi = phi0.copy()
    for i in range(niter):
        inside = I[phi < 0]
        outside = I[phi > 0]
        u = np.mean(inside) if inside.size > 0 else 0
        v = np.mean(outside) if outside.size > 0 else 0
        K = curvature(phi)
        phi += tau * lambda1 * (I - u)**2 * upwind_entropy_norm(phi, 1)
        phi -= tau * lambda2 * (I - v)**2 * upwind_entropy_norm(phi, -1)
        phi -= tau * mu * K
        phi += tau * nu * upwind_entropy_norm(phi, 1)

        int_inside = (I - u)**2
        int_outside = (I - v)**2
        contour_len = np.abs(np.sum(np.gradient(phi)))
        E[i] = (lambda1 * np.sum(int_inside[phi < 0]) +
                lambda2 * np.sum(int_outside[phi > 0]) +
                mu * contour_len +
                nu * np.sum(int_inside[phi < 0]))
        print("contour_len: ", contour_len)
    return phi, niter, u, v, E

def main():
    # Parameters
    n = 300
    addNoise = False
    filename = 'brain.png'
    I = Image.open(filename).convert('L')
    I = I.resize((n, n))
    I = np.array(I, dtype=np.float64)
    if addNoise:
        I = gaussian_filter(I, sigma=1)
    f0 = I / np.max(I)
    
    # Initial level set: single circle
    Y, X = np.meshgrid(np.arange(n), np.arange(n))
    phi0 = np.full((n, n), np.inf)
    r = 0.1 * n
    c = (n / 2, n / 2)
    phi0 = np.minimum(phi0, np.sqrt((X - c[0])**2 + (Y - c[1])**2) - r)

    # Time and energy parameters
    deltaT = 0.2
    maxT = 0.8
    lambda1 = 1.0
    lambda2 = 1.0
    mu = 0.5
    nu = 0.015

    # Chan-Vese computation
    phi, niter, u, v, E = chan_vese(f0, phi0, deltaT, maxT, lambda1, lambda2, mu, nu)

    # Final plots
    plt.figure()
    plt.imshow(I, cmap='gray')
    plt.contour(phi, levels=[0], colors='r')
    plt.title(f'Contour at Level-Set 0 at Iteration = {niter}')
    plt.axis('off')

    plt.figure()
    segment = np.zeros_like(phi)
    segment[phi > 0] = 1
    plt.imshow(segment, cmap='gray')
    plt.title(f'Segmented Image at Iteration = {niter}')
    plt.axis('off')

    plt.figure()
    plt.plot(np.arange(1, niter + 1), E)
    plt.xlabel('Number of Iterations')
    plt.ylabel('Energy')
    plt.title('Energy vs Iteration')
    plt.show()

if __name__ == "__main__":
    main()
\end{lstlisting}

\end{document}